\documentclass{article}
\usepackage{PRIMEarxiv}

\usepackage[utf8]{inputenc} 
\usepackage[T1]{fontenc}    
\usepackage{graphicx}
\usepackage{amsmath}
\usepackage{float}
\usepackage{tabularx}
\usepackage{booktabs}
\usepackage{amsmath}
\usepackage{amssymb}
\usepackage{amsfonts}       
\usepackage{nicefrac}       
\usepackage{microtype}      
\usepackage{lipsum}
\usepackage{fancyhdr}       
\usepackage{url}
\usepackage{hyperref}

\title{Ensemble Graph Neural Networks for Probabilistic Sea Surface Temperature Forecasting via Input Perturbations}

\author{
  \begin{minipage}[t]{0.32\textwidth}
        \centering
        Alejandro J. González-Santana \\
        {\normalfont 
        Centro de Tecnologías de la Imagen \\
        Instituto Universitario de Cibernética, Empresas y Sociedad (IUCES)\\
        University of Las Palmas de Gran Canaria, Spain\\
        \texttt{alejandro.gonzalez147@alu.ulpgc.es}}
  \end{minipage}
  \hfill  
  \And
  \begin{minipage}[t]{0.32\textwidth}
        \centering
        Giovanny A. Cuervo-Londoño\\
        {\normalfont 
        Oceanografía Física y Geofísica Aplicada \\
        Instituto Universitario en Acuicultura Sostenible y Ecosistemas Marinos (ECOAQUA)\\
        University of Las Palmas de Gran Canaria, Spain\\
        \texttt{giovanny.cuervo@ulpgc.es}}
  \end{minipage}
  \hfill
  \And
  \begin{minipage}[t]{0.32\textwidth}
        \centering
        Javier Sánchez\\
        {\normalfont 
        Centro de Tecnologías de la Imagen\\
        Instituto Universitario de Cibernética, Empresas y Sociedad (IUCES)\\
        University of Las Palmas de Gran Canaria, Spain\\
        \texttt{jsanchez@ulpgc.es}}
  \end{minipage}
}

\begin{document}

\maketitle


\begin{abstract}
Accurate regional ocean forecasting requires models that are both computationally efficient and capable of representing predictive uncertainty. This work investigates ensemble learning strategies for sea surface temperature (SST) forecasting using Graph Neural Networks (GNNs), with a focus on how input perturbation design affects forecast skill and uncertainty representation. We adapt a GNN architecture to the Canary Islands region in the North Atlantic and implement a homogeneous ensemble approach inspired by bagging, where diversity is introduced during inference by perturbing initial ocean states rather than retraining multiple models. Several noise-based ensemble generation strategies are evaluated, including Gaussian noise, Perlin noise, and fractal Perlin noise, with systematic variation of noise intensity and spatial structure. Ensemble forecasts are assessed over a 15-day horizon using deterministic metrics (RMSE and bias) and probabilistic metrics, including the Continuous Ranked Probability Score (CRPS) and the Spread–skill ratio. Results show that, while deterministic skill remains comparable to the single-model forecast, the type and structure of input perturbations strongly influence uncertainty representation, particularly at longer lead times. Ensembles generated with spatially coherent perturbations, such as low-resolution Perlin noise, achieve better calibration and lower CRPS than purely random Gaussian perturbations. These findings highlight the critical role of noise structure and scale in ensemble GNN design and demonstrate that carefully constructed input perturbations can yield well-calibrated probabilistic forecasts without additional training cost, supporting the feasibility of ensemble GNNs for operational regional ocean prediction.
\end{abstract}

\newpage

\section{Introduction}

The increasing relevance of the blue economy~\cite{novellino2025}, the accelerating impacts of climate change, and the objectives defined within the Sustainable Development Goals have intensified the demand for accurate and timely ocean forecasting systems~\cite{veitch2025}. Reliable predictions of oceanographic variables, such as sea surface temperature (SST), are essential for maritime operations, ecosystem monitoring, fisheries management, and climate-related decision-making. Traditionally, these forecasts have relied on numerical ocean models that explicitly solve the physical equations governing ocean dynamics. While physically grounded and robust, such models are computationally expensive and often restricted to large operational centers, limiting their accessibility and adaptability to regional and high-resolution applications~\cite{fox2019challenges}.

Recent advances in machine learning (ML) have introduced an alternative paradigm for geophysical prediction, enabling data-driven models to learn complex spatiotemporal relationships directly from observations and reanalysis products~\cite{reichstein2019deep}. Leveraging modern hardware accelerators, ML-based models offer orders-of-magnitude faster inference compared to numerical solvers, making them particularly attractive for operational and near–real-time forecasting. Although many state-of-the-art ML climate models operate at global scales, recent work has demonstrated their growing potential for regional applications, where higher spatial resolution is required to capture localized ocean processes such as coastal upwelling and mesoscale variability.

Within this context, Graph Neural Networks (GNNs) have emerged as a powerful framework for modeling geophysical systems defined over irregular spatial domains~\cite{battaglia2018relational}. By representing spatial locations as nodes and their physical relationships as edges, GNNs naturally handle complex geometries such as coastlines and bathymetry, which are difficult to model using regular grids. Architectures such as GraphCast~\cite{lam2023learning} and its regional adaptations~\cite{oskarsson2023graph} have demonstrated that GNNs can achieve competitive or superior performance to traditional numerical models in medium-range forecasting while maintaining high computational efficiency. Recent  works~\cite{holmberg2024regional,cuervo2025deeplearning,cuervo2025deeplearning2} extend this approach to regional ocean prediction through a hierarchical encoder–processor–decoder GNN tailored to oceanographic data.

Despite these advances, a major limitation of most deep learning–based forecasting systems is their deterministic nature, which hampers their ability to represent forecast uncertainty---an essential component for operational oceanography and climate services~\cite{gneiting2014probabilistic}---. Ensemble forecasting~\cite{zhou2002ensembling} is the standard approach for uncertainty quantification in numerical weather and ocean prediction, typically achieved through perturbed initial conditions or stochastic parameterizations. Recently, ensemble methodologies have also been adopted in machine learning–based forecasting systems, including AIFS ENS~\cite{AIFS-ENS}, GenCast~\cite{price2023gencast}, NeuralGCM~\cite{kochkov2024neuralgcm}, and Pangu-Weather~\cite{bi2023accurate}, demonstrating that ML ensembles can provide calibrated probabilistic forecasts at a fraction of the computational cost of traditional ensemble systems.

Ensemble learning relies fundamentally on diversity among its members: individual forecasts must differ sufficiently so that their errors are weakly correlated, allowing aggregation to improve robustness and reliability~\cite{dietterich2000ensemble,kuncheva2003measures}. In the context of deep learning, diversity can be introduced through heterogeneous architectures, multiple independently trained models, or perturbations applied to inputs, parameters, or latent representations. However, training multiple high-capacity models is often computationally prohibitive, particularly for regional GNNs operating on high-resolution meshes.

This work explores a computationally efficient ensemble strategy for regional ocean forecasting using GNNs, implemented through input perturbation during inference rather than repeated training. Building on SeaCast~\cite{holmberg2024regional}, we adapt the model to the Canary Islands region in the North Atlantic---an area characterized by strong upwelling dynamics and high spatiotemporal variability---and investigate how different noise-based perturbation strategies affect ensemble performance. Specifically, we compare Gaussian noise and spatially structured perturbations based on Perlin and fractal Perlin noise, systematically analyzing the influence of noise intensity and spatial resolution.

The ensemble forecasts are evaluated using both deterministic metrics (RMSE and bias) and probabilistic metrics from WeatherBench~\cite{rasp2020weatherbench}, including the Continuous Ranked Probability Score (CRPS) and the spread–skill ratio. These metrics allow us to assess not only forecast accuracy but also ensemble calibration and uncertainty representation across multiple lead times. By focusing on inference-time perturbations and ensemble design choices, this study aims to clarify how uncertainty can be effectively represented in GNN-based regional ocean forecasts under realistic computational constraints.

In contrast to recent generative or multi-model ensemble approaches, our focus is deliberately on lightweight ensemble construction suitable for regional, high-resolution GNN deployments under constrained computational budgets. Rather than retraining multiple networks or introducing stochasticity during training, we investigate how carefully designed perturbations of the initial ocean state at inference time can induce meaningful forecast diversity. This design choice enables rapid ensemble generation from a single trained model while preserving operational feasibility. 

The main contributions of this work are threefold: First, we present an efficient ensemble framework for regional SST forecasting based on inference-time input perturbations applied to a hierarchical GNN; Second, we provide a detailed comparison of unstructured (Gaussian) versus spatially coherent (Perlin and fractal Perlin) noise, demonstrating that the latter yields more reliable uncertainty estimates at longer lead times; Third, we offer empirical guidance on noise scale and structure for ensemble GNN design in regional ocean applications. 

The remainder of the paper is organized as follows: Section~\ref{se:related_work} reviews related work on ML-based geophysical forecasting and ensemble methods; Section~\ref{se:methods_data} describes the dataset, GNN architecture, and ensemble generation strategies; Section~\ref{se:experimental_setup} outlines the experimental setup and evaluation metrics; Section~\ref{se:results} presents the results; and Section~\ref{se:discussion} discuss the implications of our findings and conclude with directions for future research.

\section{Related Work}
\label{se:related_work}
Ensemble forecasting is central to uncertainty quantification in numerical weather prediction (NWP), where diversity is typically introduced through perturbed initial conditions and stochastic parameterizations. As ML models increasingly complement or replace traditional solvers in atmospheric and ocean forecasting, ensemble methodologies have been adapted to neural architectures~\cite{lakshminarayanan2017deepensembles} to provide probabilistic predictions. Recent work in ML-based geophysical forecasting can be organized into four principal ensemble paradigms: (i) independently trained multi-model ensembles, (ii) generative probabilistic models, (iii) hybrid physics–ML systems with stochastic components, and (iv) inference-time perturbation ensembles.

Operational AI systems such as AIFS ENS~\cite{AIFS-ENS} follow the classical ensemble paradigm by training or fine-tuning multiple model instances. Diversity arises from independent optimization trajectories, initialization differences, or data subsampling, consistent with ensemble theory \cite{dietterich2000ensemble,sagi2018survey,yang2023survey}. This strategy benefits from strong variance reduction and often yields well-calibrated forecasts. However, it requires substantial computational resources, limiting scalability for regional, high-resolution domains where training multiple large neural models is prohibitive.

A second paradigm is represented by diffusion-based systems such as GenCast~\cite{price2023gencast}. These models learn the conditional distribution of future states and directly generate multiple coherent forecast trajectories. By optimizing distributional objectives, they achieve competitive CRPS performance relative to operational ensemble systems. Nevertheless, generative approaches~\cite{couairon2024archesweather,bodnar2025aurora,medina2025leveraging} introduce considerable training complexity, high computational costs, and reduced interpretability, as uncertainty is encoded implicitly in latent representations rather than explicitly through physically motivated perturbations.

NeuralGCM \cite{kochkov2024neuralgcm} exemplifies a hybrid strategy in which neural components are embedded within a physically inspired dynamical core. Stochastic perturbations are applied to learned tendencies, and probabilistic losses such as CRPS guide training. This approach enhances physical consistency and uncertainty representation but increases architectural complexity and coupling between model components, complicating adaptation to irregular regional meshes and ocean-specific geometries.

Recent work highlights the importance of spatially coherent perturbations. Pangu-Weather~\cite{bi2023accurate} demonstrated that structured Perlin noise ~\cite{perlin1985image} applied to initial states can generate meaningful ensemble diversity while preserving spatial smoothness. Compared to spatially independent Gaussian perturbations, structured noise better respects geophysical correlation scales, reducing artificial high-frequency artifacts. However, prior studies have primarily focused on global atmospheric forecasting and have not systematically examined how perturbation resolution and scale influence probabilistic skill, particularly in regional ocean contexts.

In parallel, Graph Neural Networks (GNNs)~\cite{bronstein2021gnn,zhou2020comprehensive} have emerged as powerful tools for geophysical prediction~\cite{battaglia2018relational,keisler2022forecasting} on irregular domains~\cite{reyes2025adaptive,cuervo2025voronoi}. Systems such as GraphCast~\cite{lam2023learning} demonstrate that encoder–processor–decoder GNN architectures achieve state-of-the-art medium-range forecasting skill with high computational efficiency. Yet these models are predominantly deterministic, and ensemble strategies for GNN-based geophysical forecasting remain underexplored. Broader GNN ensemble approaches introduce diversity via architectural variation or subgraph sampling \cite{wei2023gnnensemble,wong2023ensemble}, but these strategies typically require retraining and are not tailored to probabilistic forecasting metrics.

Existing ensemble paradigms reveal a trade-off: generative and multi-model systems offer strong probabilistic calibration but incur high computational cost, while simple perturbation schemes are inexpensive but may lack physical realism. This work occupies a computationally efficient middle ground. We investigate homogeneous ensembles for hierarchical GNNs~\cite{oskarsson2023graph,oskarsson2024probabilistic} in regional sea surface temperature forecasting~\cite{holmberg2024regional,cuervo2025forecasting,cuervo2025deeplearning2}, introducing diversity exclusively at inference time through controlled perturbations of the initial ocean state. Unlike prior studies, we provide a systematic comparison between unstructured Gaussian noise and spatially coherent Perlin-based perturbations, explicitly analyzing the impact of noise structure and resolution on CRPS and spread–skill ratio \cite{zamo2018estimation,rasp2020weatherbench,rasp2023weatherbench2}. Our results demonstrate that spatial coherence, rather than perturbation complexity, is the dominant factor in achieving well-calibrated medium-range forecasts.

By focusing on structured inference-time perturbations within a regional GNN framework, this study establishes a lightweight and operationally feasible pathway toward probabilistic ocean forecasting, bridging deterministic neural models and resource-intensive generative ensemble systems.

\section{Methods and Data}
\label{se:methods_data}

\subsection{Dataset}
\begin{figure}[ht!]
\centering
\includegraphics[width=0.5\linewidth]{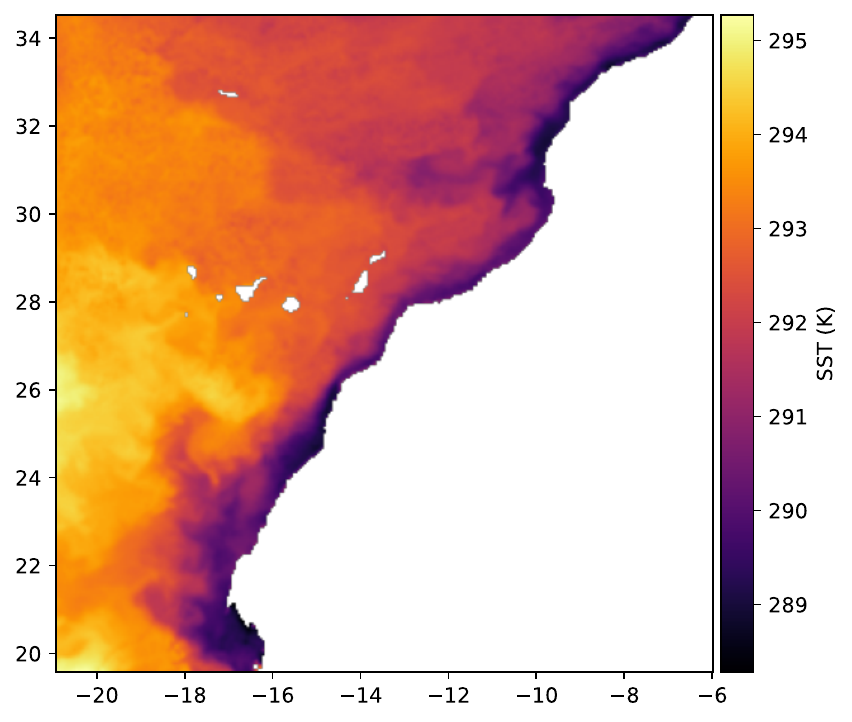}
\caption{Sea Surface Temperature (SST) data in Kelvin for January 1, 2018, in the north-west African coast.}
\label{fig:sst_ex_1}
\end{figure}

This study relies on three primary data types: oceanographic, atmospheric forcing, and bathymetry. Oceanographic data were obtained from the Copernicus Marine Service (CMEMS), specifically the European North-West Shelf / Iberia-Biscay-Irish Seas High-Resolution L4 Sea Surface Temperature Reprocessed product. This Level 4 reanalysis merges satellite observations with quality control and gap-filling, providing consistent daily SST data from January 1, 1982, to December 31, 2023, at 0.05°×0.05° resolution over the North Atlantic, Bay of Biscay, Irish Sea, and part of the western Mediterranean. The spatial domain is bounded by the coordinates listed in Table~\ref{tab:coordinates}. SST is measured at 20\,cm depth, with an accompanying error field (not used for model training). Data were processed into daily matrices. Figure \ref{fig:sst_ex_1} shows SST for January 1, 2018, in our domain. 

\begin{table}[ht!]
\caption{Coordinates delimiting the area of study.}
\label{tab:coordinates}
\centering
\begin{tabular}{|l|c|}
\hline
\textbf{Parameter} & \textbf{Value} \\
\hline
Maximum latitude  & 34.525º \\
Minimum latitude  & 19.55º \\
Maximum longitude & -5.975º \\
Minimum longitude & -20.97º \\
\hline
\end{tabular}
\end{table}

\begin{figure}[ht!]
\centering
\includegraphics[width=1\linewidth]{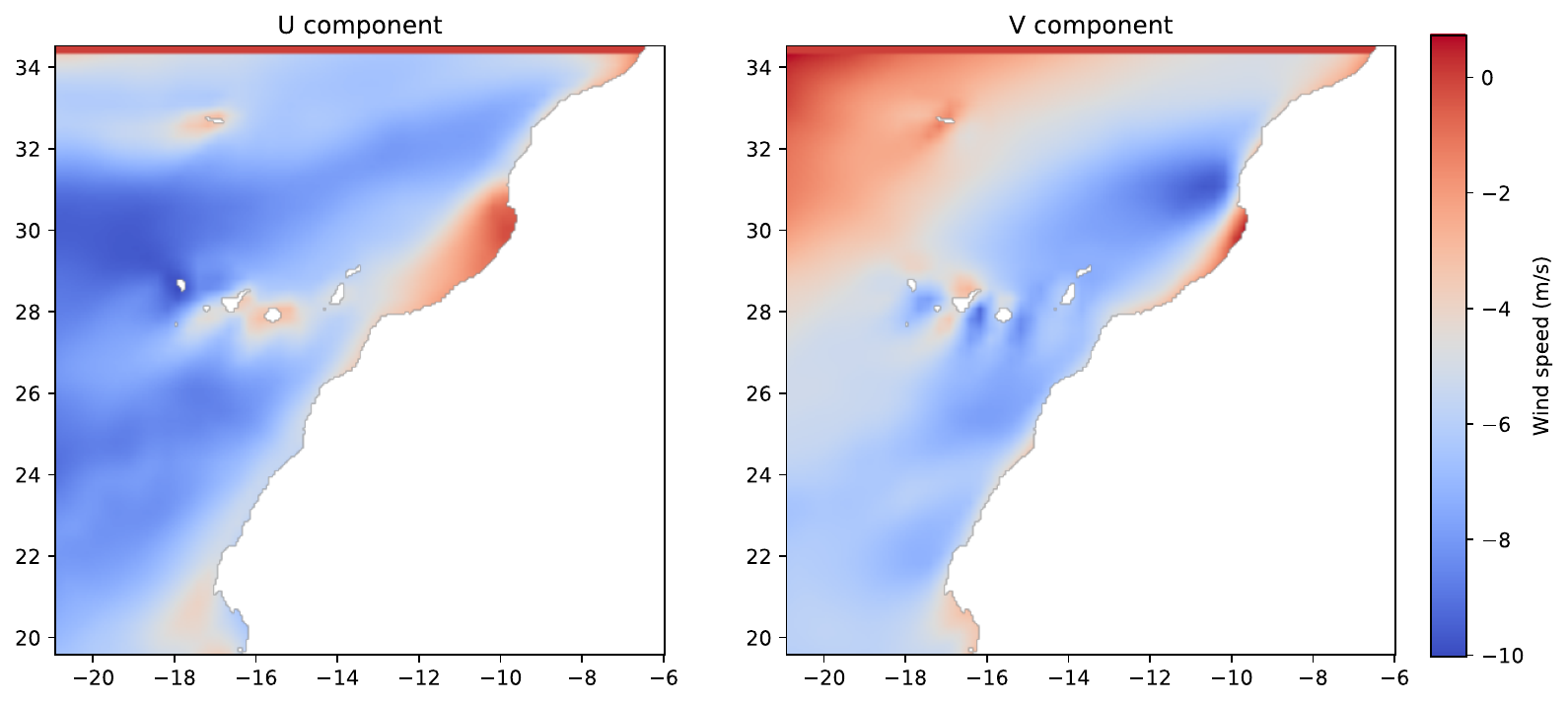}
\caption{Wind components ($u$ and $v$) at 10 meters above the surface for January 1, 2018.}
\label{fig:comp_u_v}
\end{figure}

Atmospheric forcing data were obtained from the Copernicus Climate Data Store (C3S) ERA5 hourly data on single levels (1940–present). Two variables were selected: the east–west ($u10$) and north–south ($v10$) components of 10\,m wind speed. Data were aggregated to daily means and interpolated to the CMEMS SST grid, stored as daily arrays of ocean points. Figure \ref{fig:comp_u_v} shows $u$ and $v$ winds for January 1, 2018.

\begin{figure}[ht!]
\centering
\includegraphics[width=0.5\linewidth]{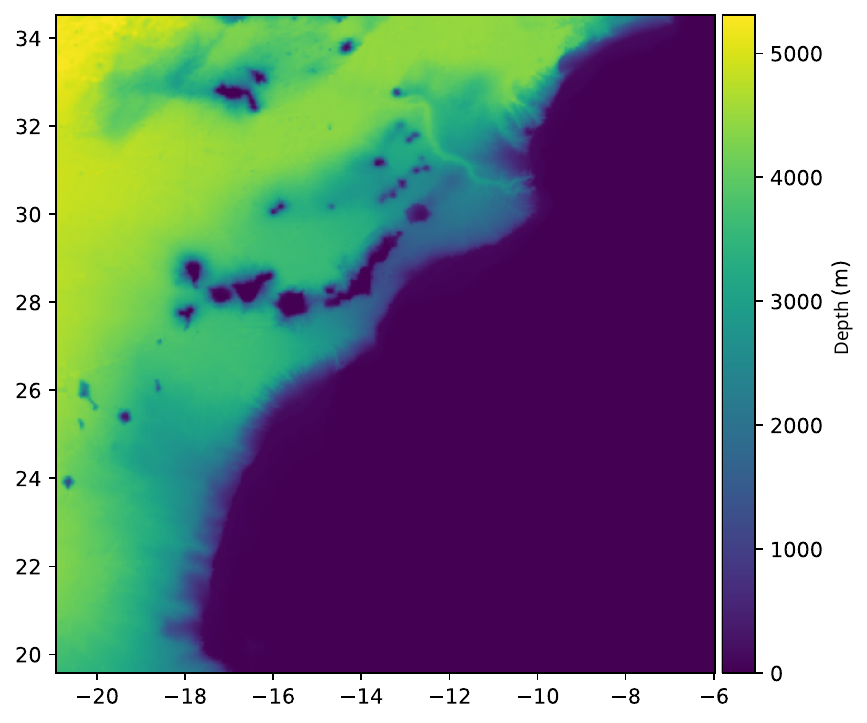}
\caption{Processed bathymetry downloaded from ETOPO.}
\label{fig:bat}
\end{figure}

Bathymetry data were obtained from NOAA’s ETOPO Global Relief Model 2022, providing ocean depth and land elevation at 0.0083° resolution. Depths were extracted for the study area, resampled to the SST grid, and land values were set to zero. Figure \ref{fig:bat} shows the processed bathymetry.

\subsection{Graph Neural Network: Adapting to the North-Atlantic Subregion}

SeaCast is a Graph Neural Network (GNN) designed for high-resolution, medium-term ocean predictions. The model leverages the complex geometry of the oceans through an autoregressive approach, combining historical oceanographic states, external forcings, and static information to predict future states. Originally developed for the Mediterranean Sea, it has been adapted in this study to the North Atlantic region.

It formalizes the prediction problem using temporal windows. Let the historical states of the oceanographic variables be denoted as
\[
X^{-h:0} = (X^{-h}, \ldots, X^{0}),
\]
where $X^{-h}$ represents the state $h$ days in the past and $X^{0}$ the current state. The future prediction window is
\[
X^{1:T} = (X^{1}, \ldots, X^{T}),
\]
where $T$ denotes the prediction horizon in days. Forcing factors are defined analogously as
\[
F^{1:T} = (F^{1}, \ldots, F^{T}).
\]

\begin{figure}[ht!]    
    \centering    
    \includegraphics[width=0.9\linewidth]{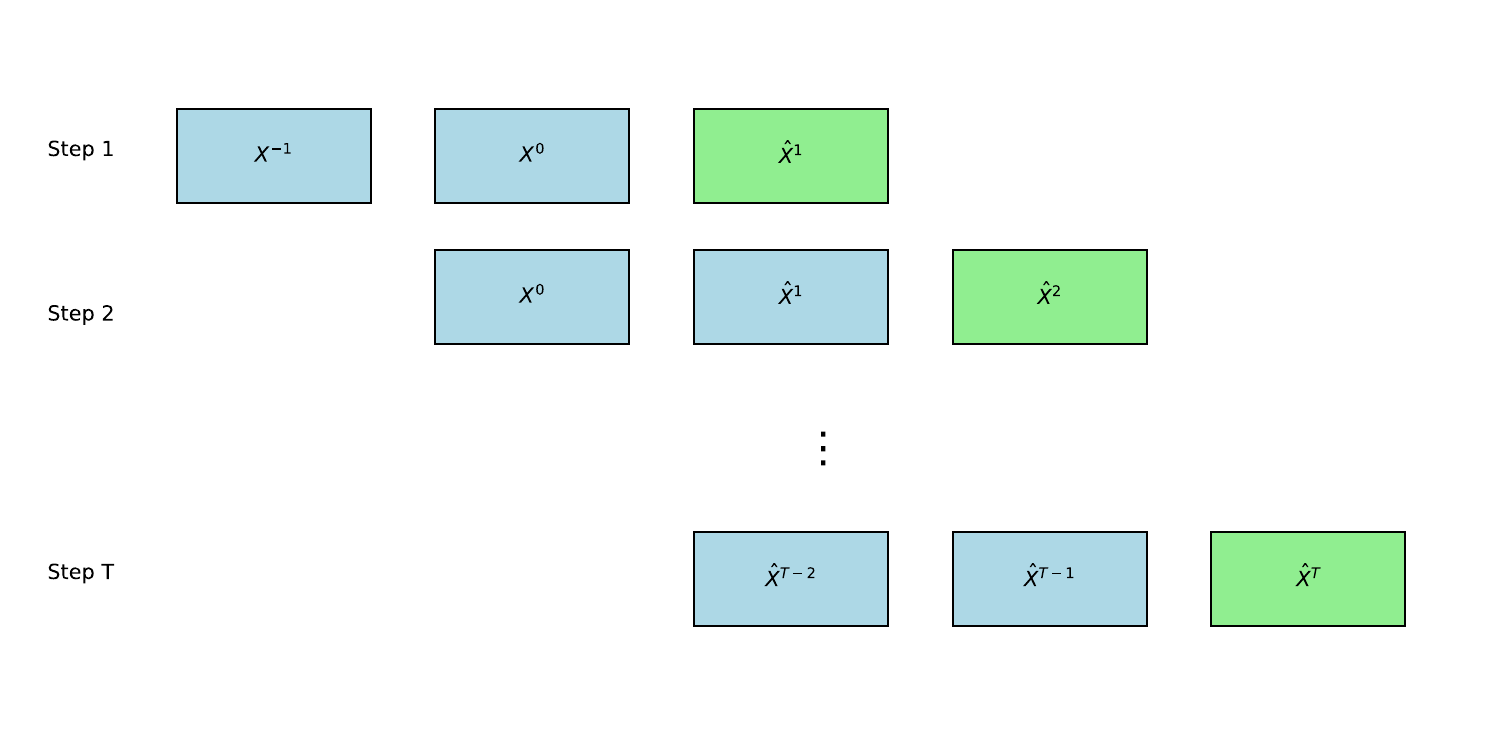}    
    \caption{Diagram of autoregressive operation.}    
    \label{fig:esqautoregresivas}
\end{figure}

The prediction relies on four components: initial and target states, forcings, and static data. The initial states consist of the previous day and the current value, while the target states correspond to the oceanographic variables over the $T$ forecasted days. In this study, we set $T=1$ for training and validation, and $T=15$ for testing. During training, the model only requires one day of forcing information, as it is trained to predict one time step ahead. Thanks to its autoregressive design, predictions can be extended to any length $T$ by sequentially generating states $\hat{X}^{1:T}$, starting from $X^0$. Figure \ref{fig:esqautoregresivas} illustrates the autoregressive process.

The SeaCast GNN uses an encoder–processor–decoder architecture applied to a hierarchical mesh graph~\cite{oskarsson2023graph}. Each stage uses an independent GNN, known as an interaction network~\cite{battaglia2016interaction}, designed to capture relationships between nodes and external influences. Latent representations are updated through message passing, MLPs with Swish activation, and layer normalization.  

\paragraph{Hierarchical Mesh Graph.} The mesh graph represents the spatial domain at multiple resolutions. Each node connects to its eight nearest neighbors (except boundary nodes) and hierarchically communicates with lower-resolution nodes. This design allows the network to capture both local and regional patterns efficiently. 

\paragraph{Encoder.} The encoder maps the inputs from the latitude-longitude grid onto the hierarchical mesh graph. A bipartite graph connects the grid to the mesh nodes, propagating information sequentially through the hierarchy. At the end of the encoding phase, all nodes contain enriched latent representations suitable for processing.

\paragraph{Processor.} In the processing stage, the latent representations are updated sequentially across mesh levels. Multiple processing layers can be used to deepen representation learning, allowing the model to capture complex spatiotemporal patterns. The model predicts the residual change as:
\begin{equation}
\hat{X}^t_r = f\left(X^{t-2:t-1}, F^t\right)_r = X^{t-1}_r + \text{MLP}^{\text{pred}}(v^G_r),
\label{eq:predfinal}
\end{equation}
where $v^G_r$ represents the latent graph representation of grid cell $r$ obtained from the mesh-to-grid connections. This residual formulation improves training stability and prediction accuracy.

\paragraph{Decoder.} The decoder maps the updated representations back from the hierarchical mesh to the latitude-longitude grid. A downward sweep through the hierarchy generates the predicted states for each grid cell.

To adapt the model to the North Atlantic, we made the following changes:

\begin{itemize}
    \item Input constants were modified to accommodate new data sources, grid dimensions, and the number of oceanographic and atmospheric variables.
    \item The mesh was adapted to the north-west African coast, including the Canary Islands.  
    \item Training was run on a single GPU instead of the original 32-GPU parallel configuration.
    \item A noise module was introduced to generate diverse ensemble predictions.
    \item Evaluation tools from WeatherBench were modified to handle the custom data format.
\end{itemize}

These modifications allow efficient operation in the North Atlantic while maintaining the core GNN architecture and autoregressive prediction capabilities. The model was trained to minimize the Mean Squared Error (MSE), considering the sequence of states generated by autoregressive steps, also known as rollout. The loss function is given by:
\begin{equation}
L = \frac{1}{T_{\text{rollout}}} \sum_{t=1}^{T_{\text{rollout}}} \sum_{i=1}^{C} \sum_{l=1}^{L_i}\frac{1}{|G_{l}|} \sum_{v \in G_{l(i)}} a_v\lambda_i  \left( \hat{X}_{v,i}^{t} - X_{v,i}^{t} \right)^2 \text{,}
\label{eq:funcperdida}
\end{equation}
where $T_\text{rollout}$ is the number of autoregressive steps,  $C$ is the total number of variables, $L_i$ is the number of depth levels of feature $i$, and $G_l$ represents the number of grid points in the ocean at level $l$. The term $a_v$ represents the weighting according to cell size, calculated as the cosine of latitude for each cell $v$ and depth level $l$. Finally, $\lambda_i$ is the inverse of the variance of the estimates for variable $i$ \cite{holmberg2024regional}.

\subsection{Ensemble Method}
Our ensemble learning strategy combines multiple models to generate a prediction during inference. This approach, inspired by the \emph{wisdom of the crowd}, relies on the expectation that individual model errors will partially cancel out, yielding more accurate predictions than a single model. The outputs are combined using an aggregation function~\cite{sagi2018survey}.

For an ensemble to outperform an individual model, base models must be better than random, and their errors should be as independent as possible \cite{dietterich2000ensemble}. Diversity among models is crucial: heterogeneous ensembles use different architectures (e.g., Stacking), while homogeneous ensembles use identical architectures with variations in inputs or parameters.

In this study, we adopted a bagging-inspired homogeneous ensemble strategy. Instead of retraining multiple models on bootstrapped datasets, a single GNN model is employed, and diversity is introduced at inference time by perturbing the oceanographic input data with noise. Predictions generated from perturbed initial states are aggregated through their mean to produce the final forecast. The types of noise and their characteristics are described in the following subsections.

\subsubsection{Gaussian}
The simplest approach to introducing variability into the data is by adding random perturbations drawn from a normal distribution. Gaussian noise is widely used due to its prevalence in natural processes and favorable statistical properties. The distribution is given by:

\begin{equation}
p(x) = \frac{1}{\sqrt{2\pi\sigma^2}} \, e^{-\frac{(x - \mu)^2}{2\sigma^2}},
\label{eq:normal}
\end{equation}
where $\mu$ denotes the mean and $\sigma^2$ the variance. Gaussian noise has been employed in several state-of-the-art climate modeling frameworks included in performance benchmarks~\cite{rasp2023weatherbench2}, such as ArchesWeatherGen~\cite{couairon2024archesweather}, NeuralGCM~\cite{kochkov2024neuralgcm}, FuXi~\cite{chen2023fuxi}, and FourCastNet~\cite{pathak2022fourcastnet}.

\begin{figure}[ht!]
\centering
\includegraphics[width=1\linewidth]{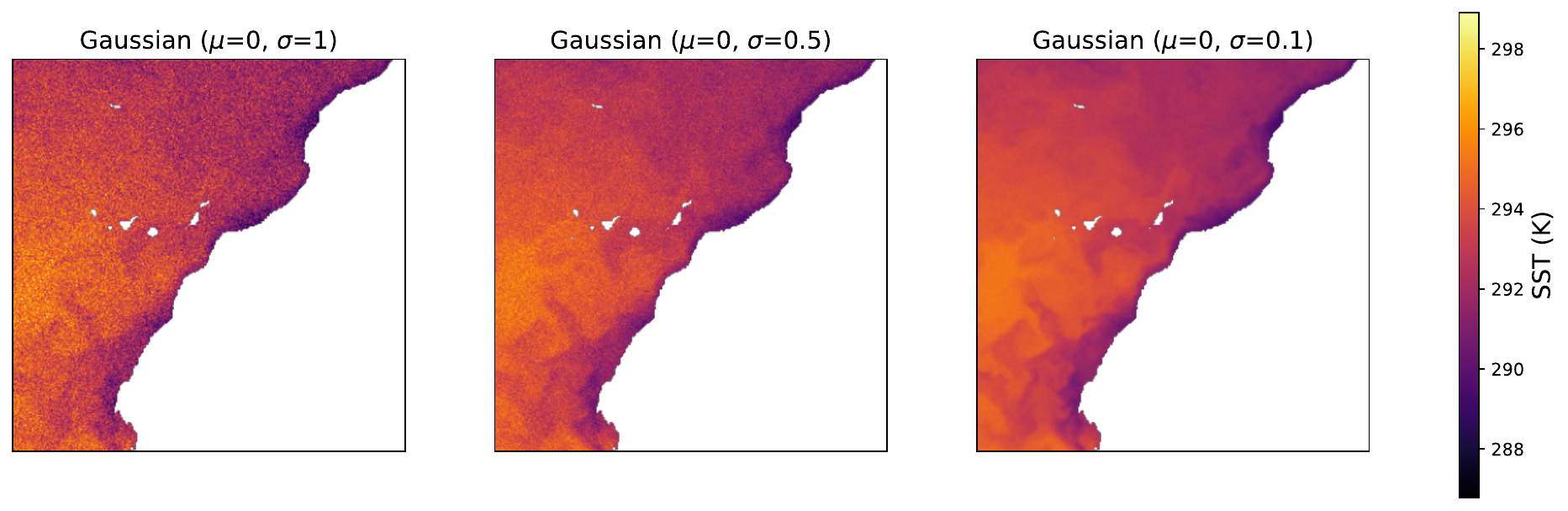}
\caption{Examples of Gaussian noise added to SST, generated with different standard deviations.}
\label{fig:gaussexample}
\end{figure}

Figure~\ref{fig:gaussexample} illustrates the effect of varying the standard deviation of the sampling distribution on noise intensity, assuming zero mean in all cases. A standard deviation of 1 yields perturbations that can reach values close to 4 in the distribution tails, whereas a standard deviation of 0.5 produces noticeably lower-amplitude noise due to reduced dispersion. When the standard deviation is set to 0.1, the perturbations become barely perceptible at the shared visualization scale. Despite these differences in magnitude, all realizations exhibit a spatially uncorrelated structure, resulting in uniformly scattered patterns across the domain.

\subsubsection{Perlin Noise}

In the context of ensemble construction, Perlin noise~\cite{perlin1985image} offers a structured approach to promoting diversity while preserving spatial correlations. In contrast to Gaussian noise, which introduces spatially independent perturbations, Perlin noise generates smooth variations with controllable correlation lengths determined by the selected spatial resolution. This property enables the creation of ensemble members that differ in a physically consistent manner, maintaining large-scale coherence while introducing localized variability. Consequently, this type of noise facilitates diversification without injecting spurious high-frequency artifacts, making it particularly suitable for perturbing geophysical variables characterized by strong spatial dependencies. Perlin noise has been notably applied in Pangu-Weather~\cite{bi2023accurate} to generate random perturbations .

To generate the noise, a three-dimensional grid is first considered, whose points have integer coordinates. These points, which act as the vertices of the grid cubes, are assigned pseudo-random values and gradient vectors through a hash function $H$, which produces four mutually independent real values. For a point with coordinates $(x, y, z)$, this assignment is represented by the following equation:

\begin{equation}
[a, b, c, d] = H(x, y, z),
\label{eq:perlinassing}
\end{equation}
where $a$, $b$ and $c$ represent the gradient and $d$ the value of the function at the point $(x, y, z)$.

If the evaluated point $(x, y, z)$ coincides with a grid point, then the noise value equals $d$. Otherwise, if the point lies within the volume bounded by the vertices of a cube, the noise is computed through smooth interpolation (e.g., a cubic polynomial) using the values and gradients of the surrounding vertices. This interpolation is performed first along the $x$-axis (between edges), then along the $y$-axis (across cube faces), and finally along the $z$-axis (between planes). This procedure allows the generation of noise with a spatial component.

\begin{figure}[ht!]
    \centering
    \includegraphics[width=0.7\linewidth]{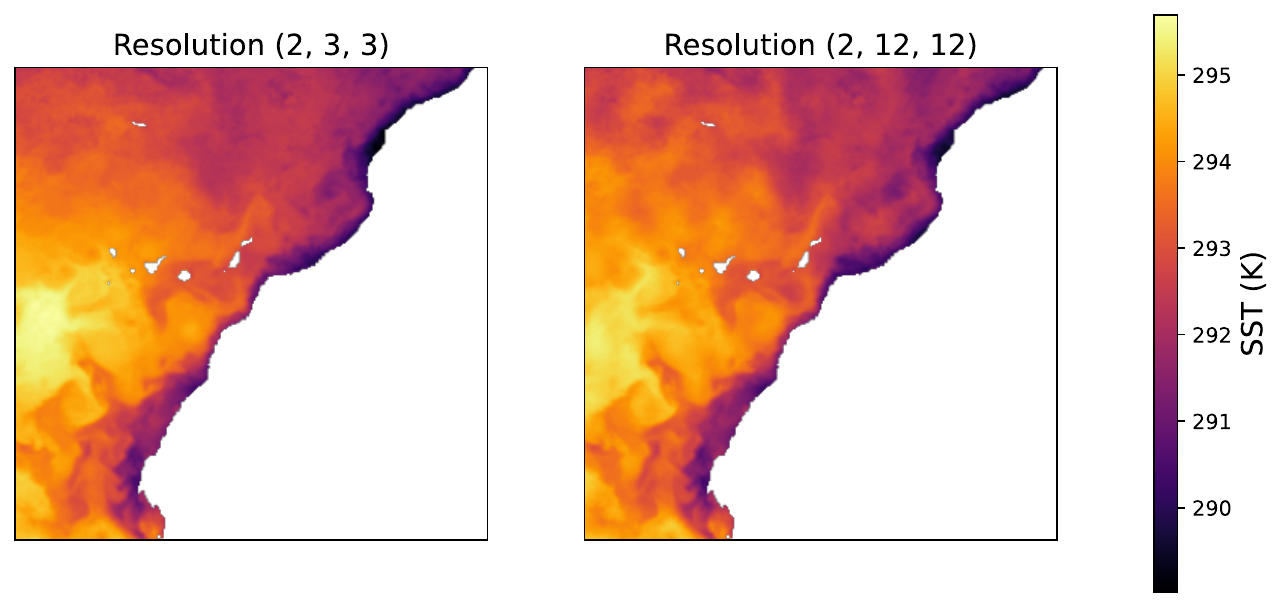}
    \caption{Examples of Perlin noise added to SST, generated with different spatial resolutions.}
    \label{fig:exampleperlin}
\end{figure}

Figure \ref{fig:exampleperlin} shows two different Perlin noise configurations that differ in the spatial resolution applied. In the case of higher spatial resolution on the latitude and longitude axes (2, 12, 12), a greater number of noise patterns can be observed, while in the low-resolution configuration (2, 3, 3), the patterns are larger and it is clearer to see how the interpolation smooths the transitions within each cell. Another important feature is that the higher resolution noise has a slightly higher intensity.

\subsubsection{Fractal Perlin Noise}
Another alternative for generating noise is the fractal Perlin noise. This function is distinguished by the use of three additional parameters. The first is \textit{octaves}, which represent the number of iterations of the noise. Then there is \textit{persistence}, which acts as the scaling factor for the noise amplitude between consecutive octaves. Finally, \textit{lacunarity} is the factor of increase in frequency between two octaves. Its equation is given as:

\begin{equation}
F(x, y, z) = \alpha \cdot \sum_{i=0}^{O-1} a_i \cdot H(x, y, z; f_i) \text{,}
\label{eq:fractalperlin}
\end{equation}
where $\alpha$ represents the noise scale, $O$ represents the number of octaves, $a_i$ is the noise amplitude, and $fi$ is the frequency that modifies the resolution. The lacunarity, persistence, and resolution evolve as follows:

\begin{equation}
resolution_{i} =  f_i \cdot resolution_{0} \text{;}  \quad f_i = lacunarity \cdot f_{i-1} \text{;} \quad a_i = persistence \cdot a_{i-1} \text{.}
\label{eq:evofractal}
\end{equation}

The noise resolution applied in each octave must always be an exact divisor of the dimensions of the desired noise shape.

Figure \ref{fig:exampleperlin} shows two different configurations of fractal Perlin noise that differ only in the spatial resolution applied. Three octaves are used in the noise iteration, with a persistence of 0.5, which means that the intensity decreases progressively with each iteration. The lacunarity is two, which means that as the octaves progress, smaller and smaller patterns are used, resulting in more detailed and complex noise. In addition, a noise scalability coefficient ($\alpha=0.2$) is applied to the noise resulting after the last octave.

\begin{figure}[ht!]
    \centering
    \includegraphics[width=0.7\linewidth]{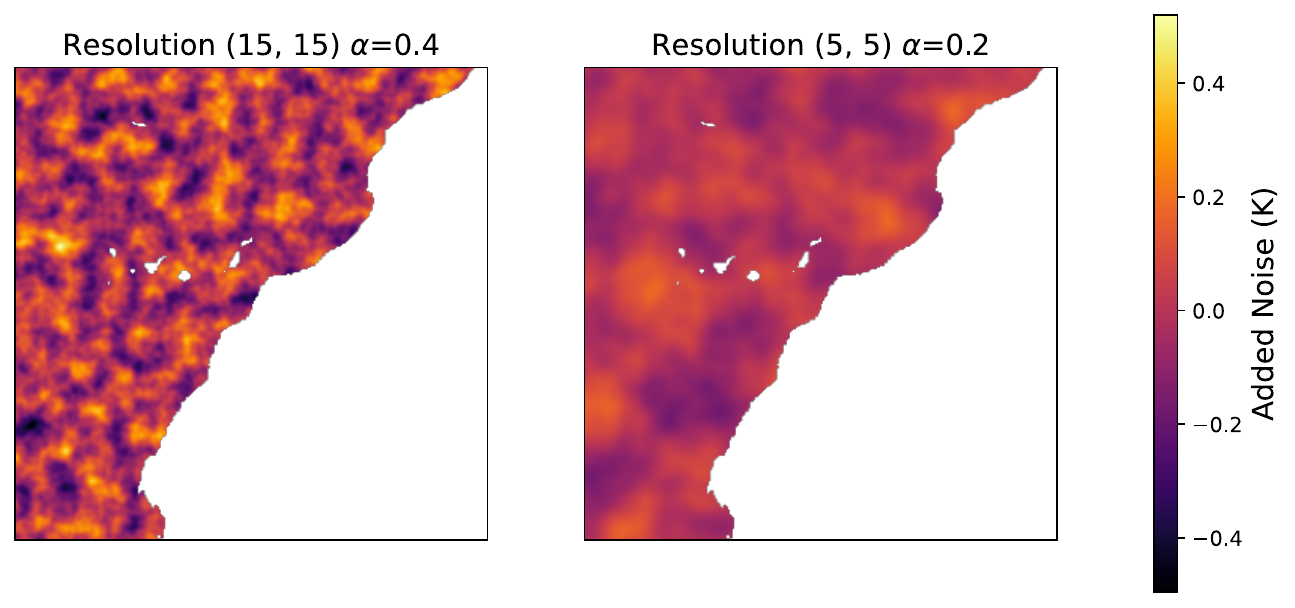}
    \caption{Examples of Perlin Fractal noise with three octaves added to the SST, generated with different spatial resolutions and the same noise scalability. The rest of parameters correspond to default values.}
    \label{fig:exampleperlinfractal}
\end{figure}

\section{Experimental Setup}
\label{se:experimental_setup}

\subsection{Evaluation Metrics}
Two sets of metrics are used in this study: probabilistic metrics to evaluate the ensemble predictions generated from noisy inputs, and classical deterministic metrics, such as RMSE or Bias, to compare predictions against a single model without noise. This section focuses on the probabilistic metrics. Table \ref{tab:notmet} summarizes the notation used throughout the metric definitions.

\begin{table}[ht!]
\caption{Notation used in the metrics \cite{rasp2023weatherbench2}.}
\label{tab:notmet}
\centering
\begin{tabular}{lll}
\toprule
\textbf{Symbol} & \textbf{Rank} & \textbf{Description} \\
\midrule
$f$ & - & Prediction \\
$o$ & - & Observation \\
$t$ & $1, \ldots, T$ & Verification time \\
$l$ & $1, \ldots, L$ & Prediction horizon \\
$i$ & $1, \ldots, I$ & Latitude index \\
$j$ & $1, \ldots, J$ & Longitude index \\
$m$ & $1, \ldots, M$ & Set member index \\
\bottomrule
\end{tabular}
\end{table}

The Continuous Ranked Probability Score (CRPS) evaluates the accuracy of ensemble predictions, balancing the error against the internal dispersion of the set. For individual members $X$ and observation $Y$, the CRPS is given by:

\begin{equation}
\text{CRPS}_\text{skill} = \mathbb{E} \lvert X-Y\rvert,
\label{eq:crpskill}
\end{equation}

\begin{equation}
\text{CRPS}_\text{spread} = \mathbb{E}[X-X'],
\label{eq:crpspread}
\end{equation}

\begin{equation}
\text{CRPS} = \text{CRPS}_\text{skill} - \frac{1}{2} \text{CRPS}_\text{spread} = \mathbb{E}\lvert X-Y\rvert - \frac{1}{2} \mathbb{E}\lvert X-X'\rvert.
\label{eq:crpsfirst}
\end{equation}

The CRPS used in this study accounts for all set members and verification times \cite{zamo2018estimation} and is calculated as:

\begin{equation}
\text{CRPS}_l := \frac{1}{T} \sum_{t}^{T} \left(
\frac{1}{M} \sum_{m=1}^{M} \left\| f^{(m)} - o \right\|_{t, l}
-
\frac{1}{2 M (M-1)} \sum_{m=1}^M \sum_{\substack{n=1 \\ n\neq m}}^M \left\| f^{(m)} - f^{(n)} \right\|_{t, l}
\right) \text{.}
\label{eq:crpsreal}
\end{equation}

The first term of the metric measures the average error of ensemble members relative to observations, and the second term provides an unbiased estimate of internal dispersion to correct for finite ensemble size.

The Spread-Skill Ratio evaluates ensemble calibration by comparing the ensemble dispersion to the RMSE of the ensemble mean. The spread is defined as:

\begin{equation}
\text{Spread}_l = \sqrt{\frac{1}{T I J} \sum_{t}^T \sum_{i}^I \sum_{j}^J\, \mathrm{var}_m \bigl(f_{t,l,i,j,m}\bigr)},
\label{eq:spread}
\end{equation}
with the variance of ensemble members given by:

\begin{equation}
\mathrm{var}_m \bigl(f_{t,l,i,j,m}\bigr) = \frac{1}{M-1} \sum_{m=1}^{M} \left( f_{t,l,i,j, m} - \bar{f}_{t,l,i,j} \right)^2.
\label{eq:varpre}
\end{equation}

The spread-skill ratio is calculated as follows:

\begin{equation}
R_l = \frac{\text{Spread}_l}{\text{RMSE}_l(\bar{f})}.
\label{eq:r_values}
\end{equation}

A value of $R_l \approx 1$ indicates good calibration, where ensemble dispersion reflects the actual prediction error. Values greater than one indicate overestimated uncertainty, while values below one indicate insufficient ensemble variability. Unlike CRPS, which balances error and dispersion, the spread-skill ratio directly measures the calibration of the ensemble~\cite{fortin2014should, wilks2011reliability}.  

In this work, a bias-corrected RMSE is used to reduce systematic errors when calculating the spread-skill ratio, providing a more reliable estimate of ensemble calibration.

\subsection{Training Procedure}
We employed a training period spanning 21 years of data with the dataset divided as follows: training data from 2003 to 2019 (17 years), validation data from 2020 to 2021 (2 years), and test data from 2022 to 2023 (2 years). This corresponds to approximately 81\% of the data for training and 9.5\% for validation and testing.

The model was trained using the AdamW optimizer with parameters $\beta_1=0.9$, $\beta_2=0.95$, and weight decay $\lambda=0.1$. The initial learning rate was set to $1\times10^{-5}$, with five warm-up epochs and a cosine decay schedule to progressively adjust the learning rate.

Network hyperparameters were selected as follows: hidden layers of 128 neurons, resulting in 128-dimensional vector representations within the GNNs and associated MLPs; four processing layers in the model processor; and a hierarchical mesh with three uniform levels at resolutions of 81, 27, and 9, respectively. Training was conducted for 150 epochs without autoregressive steps due to computational constraints.

The training environment consisted of a desktop workstation with an Intel Core i9-9900K processor (3.60 GHz, 8 cores, 16 threads), 32 GB of RAM, and an NVIDIA GeForce RTX 3060 GPU with 12 GB of VRAM, which enabled efficient handling of large datasets and accelerated computation.

Before training, static features, such as bathymetry, coordinates, and normalization statistics, were computed, and differences between consecutive states were normalized to facilitate residual learning. The structure of the hierarchical mesh and bipartite connections were generated and stored for efficient access.

During training, the model learns to predict one-step-ahead oceanographic states. For evaluation, ensemble predictions are generated by introducing diversity in the initial states through noise perturbations (Gaussian, Perlin, and Perlin fractal). The ensemble predictions are then aggregated via daily averages, enabling both probabilistic and deterministic assessment of model performance.

\subsection{Test Configurations}
Inference tests were initially performed using a single day of input to generate 15 future predictions. This preliminary phase aimed to identify noise types that performed poorly, which were discarded, while promising configurations were retained for full evaluation. Both Gaussian and Perlin-based noises were tested, with parameter settings summarized in Tables \ref{tab:gaussian_noise_tests}, \ref{tab:ruido_perlin_pruebas}, and \ref{tab:ruido_perlin_fractal_pruebas}.

Gaussian noise tests are defined by a mean ($\mu$) and standard deviation ($\sigma$). Standard deviations above 1 were avoided, as they degraded prediction quality. Table \ref{tab:gaussian_noise_tests} shows the configurations used:

\begin{table}[ht!]
\caption{Gaussian noise configurations used during the tests.}
\label{tab:gaussian_noise_tests}
\centering
\begin{tabular}{lcc}
\hline
\textbf{Noise type} & \textbf{Mean ($\mu$)} & \textbf{Standard deviation ($\sigma$)} \\
\hline
Gaussian ($\mu$=0, $\sigma$=0.1)   & 0.0  & 0.1 \\
Gaussian ($\mu$=0, $\sigma$=0.05)  & 0.0  & 0.05  \\
Gaussian ($\mu$=0, $\sigma$=0.01)  & 0.0  & 0.01  \\
\hline
\end{tabular}
\end{table}

Perlin noise tests included classic and fractal variants. Table \ref{tab:ruido_perlin_pruebas} shows classic Perlin noise configurations with different spatial resolutions, ensuring temporal consistency via the \texttt{tileable} parameter.  

\begin{table}[ht!]
\caption{Classic Perlin noise configurations used during the tests.}
\label{tab:ruido_perlin_pruebas}
\centering
\begin{tabular}{lcc}
\hline
\textbf{Noise type} & \textbf{Resolution} & \textbf{Repeatable} \\
\hline
\texttt{P\_res\_2x3x3} & (2, 3, 3)   & (T, F, F)  \\
\texttt{P\_res\_2x12x12} & (2, 12, 12) & (T, F, F) \\
\hline
\end{tabular}
\end{table}

Fractal Perlin noise was used with a spatial resolution of 15x15, three octaves, persistence of 0.5, lacunarity of 2, and a weighting factor of 0.2. Variants were generated by modifying a single parameter while keeping the recommended lacunarity and persistence, as shown in Table \ref{tab:ruido_perlin_fractal_pruebas}.

\begin{table}[ht!]
\caption{Fractal Perlin noise configurations used during the tests. The IDs correspond to: A) \texttt{PF\_res\_15x15}; B) \texttt{PF\_res\_5x5}; C) \texttt{PF\_res\_15x15\_without\_tileable}; D) \texttt{PF\_res\_15x15} ($\alpha$=0.05); and  
E) \texttt{PF\_res\_15x15} ($\alpha$=0.4).
}
\label{tab:ruido_perlin_fractal_pruebas}
\centering
\small
\begin{tabular}{ccccccc}
\hline
\textbf{ID} & \textbf{Resolution} & \textbf{Repeatable} & \textbf{Persistence} & \textbf{Octaves} & \textbf{Scale} & \textbf{Lacunarity} \\
\hline
A & (15, 15) & (F, T) & 0.5 & 3 & 0.2 & 2.0 \\
B & (5, 5) & (F, T) & 0.5 & 3 & 0.2 & 2.0 \\
C & (15, 15) & (F, F) & 0.5 & 3 & 0.2 & 2.0 \\
D & (15, 15) & (F, T) & 0.5 & 3 & 0.05 & 2.0 \\
E & (15, 15) & (F, T) & 0.5 & 3 & 0.4 & 2.0 \\
\hline
\end{tabular}
\end{table}

\section{Results}
\label{se:results}

In this section, we compare ensemble prediction with the deterministic forecast. Then we analyze the differences observed in preliminary experiments using Gaussian and Perlin noise, emphasizing how specific noise parameters influence performance when initial conditions are perturbed.

Table \ref{tab:mejora_rmse} reports the RMSE averaged over 1, 5, and 15 days, relative to the deterministic forecast without noise. None of the ensemble configurations consistently outperforms the deterministic model across all horizons, which is expected given that the model was trained under deterministic conditions.

RMSE increases with forecast horizon due to the autoregressive formulation. Ensemble predictions also exhibit higher RMSE at short lead times because the input observations are explicitly perturbed. The largest initial degradations are observed for \texttt{P$\_$res$\_$2x12x12}, \texttt{P$\_$res$\_$2x3x3}, and Gaussian ($\mu=0$, $\sigma=0.1$), reflecting their higher input variability. However, this effect decreases with lead time, and performance converges toward stable values by the final forecast day.

Overall, these results indicate that introducing diversity in the initial conditions helps compensate for forecast errors and promotes exploration of alternative future states, which becomes increasingly important as uncertainty grows in autoregressive predictions. The primary objective of this work was not to improve deterministic accuracy, but to evaluate how different ensemble strategies and input perturbations impact uncertainty characterization.

\begin{table}[ht!]
\caption{Comparison of the increase RMSE over different time horizons for each noise configuration, with respect to a deterministic prediction without noise.}
\label{tab:mejora_rmse}
\centering
\begin{tabular}{lccc}
\hline
\textbf{Configuration} & \textbf{1 day} & \textbf{5 days} & \textbf{15 days} \\ 
\hline
Deterministic (ref) & 0.109 & 0.308 & 0.586 \\ 
Gaussian ($\mu$=0, $\sigma=0.01$) & 0.55\% & 0.03\% & 0.00\% \\ 
Gaussian ($\mu$=0, $\sigma=0.1$) & 28.96\% & 5.36\% & 0.43\% \\ 
P$\_$res$\_$15x15 ($\alpha=0.05$) & 1.56\% & 0.16\% & 0.02\% \\ 
P$\_$res$\_$2x12x12 & 64.01\% & 15.90\% & 1.62\% \\ 
P$\_$res$\_$2x3x3 & 61.17\% & 14.83\% & 1.47\% \\ 
\hline
\end{tabular}

\end{table}

\begin{figure}[ht!]
    \centering
    \includegraphics[width=1\linewidth]{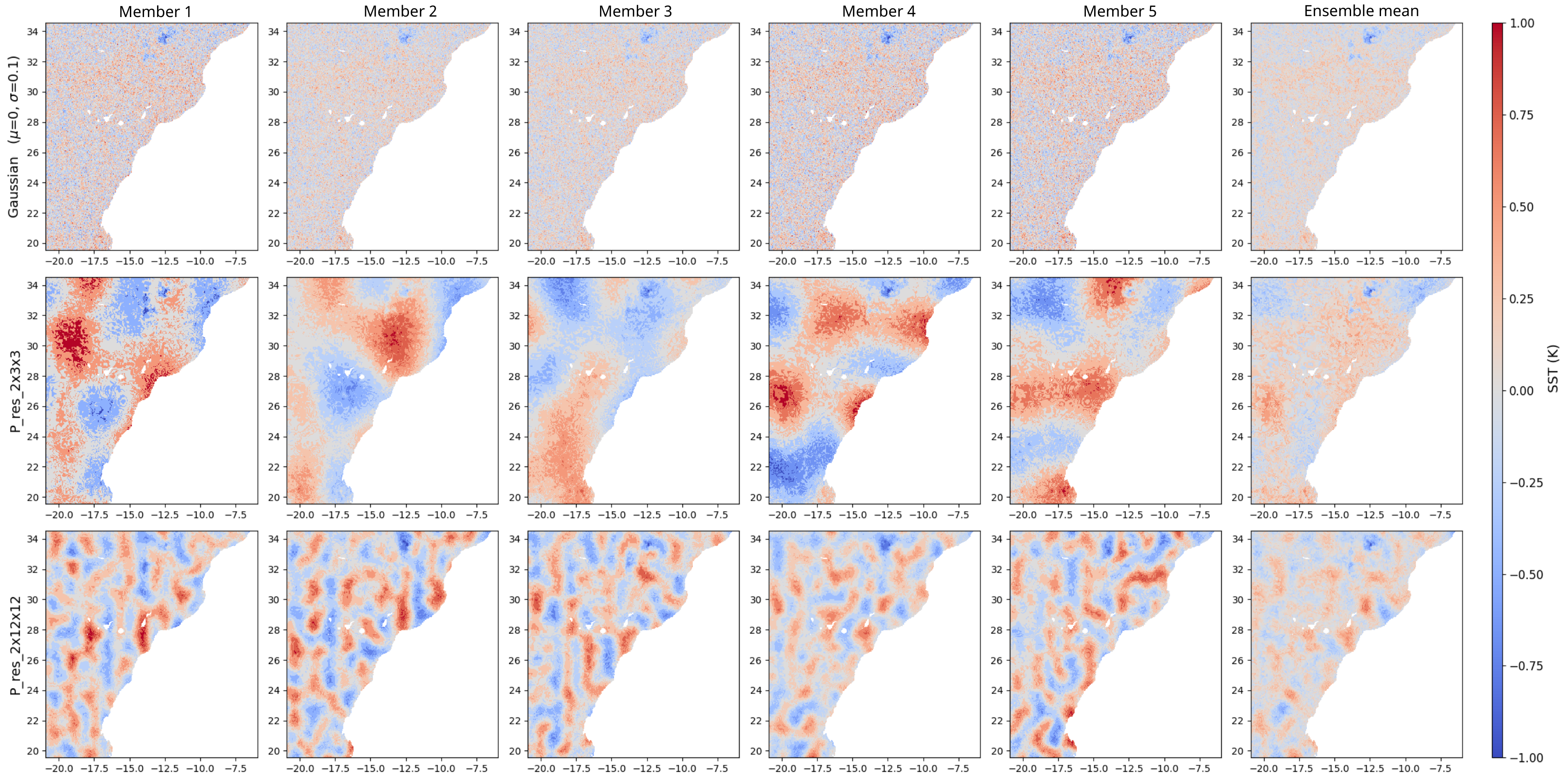}
    \caption {SST bias maps for one-day forecasts starting on 2 January 2022. Each row corresponds to a different input noise configuration, and each column represents a member of the ensemble, with the last column showing the ensemble mean.}
    \label{fig:diffensday1}
\end{figure}

Figure \ref{fig:diffensday1} shows the bias on prediction day 1 for different members of the ensemble and their mean value. The members show greater variability  while the mean of the set is more accurate.

\subsection{Comparison of Gaussian noise configurations}
\begin{figure}[ht!]
    \centering
    \includegraphics[width=0.9\linewidth]{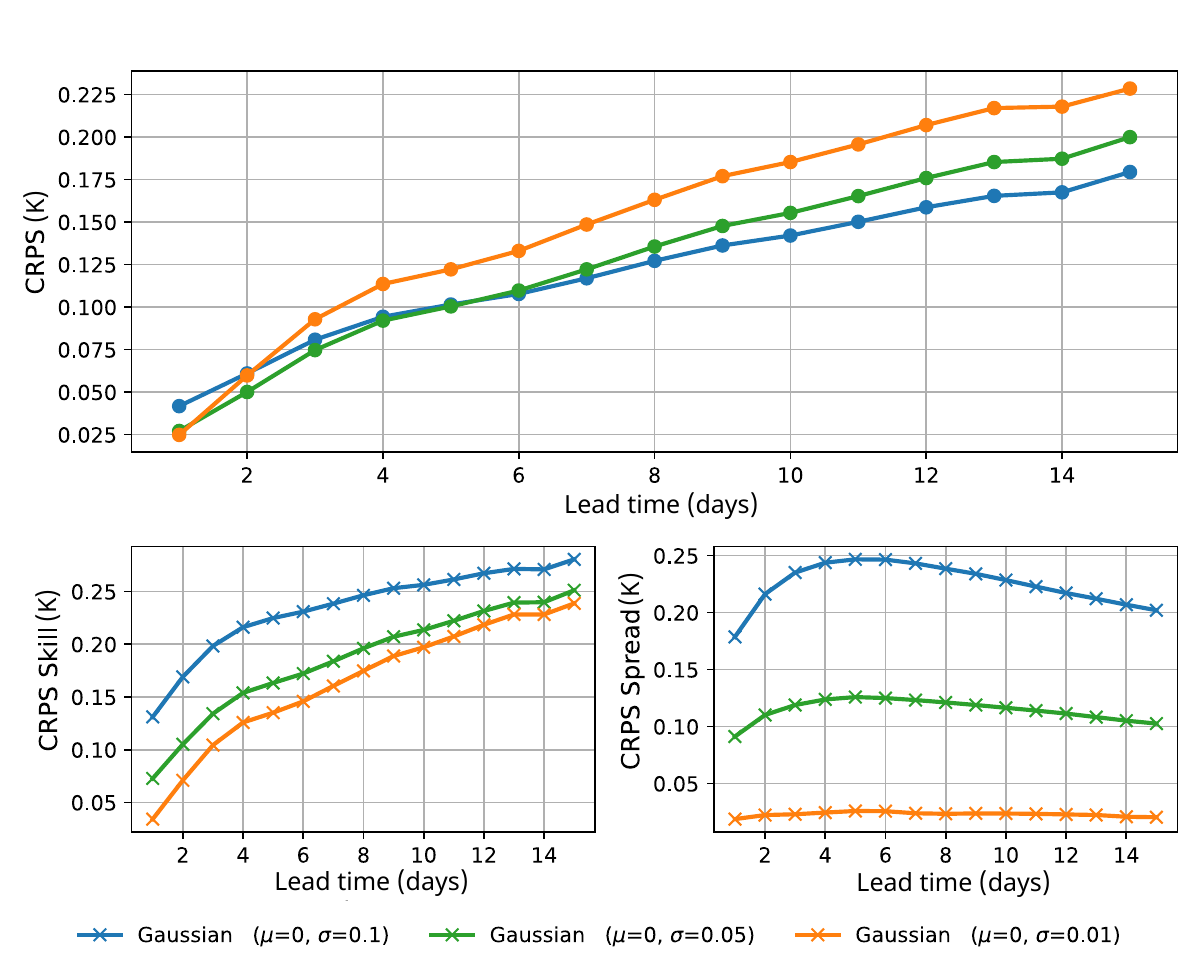}
    \caption{CRPS and its components, averaged for each prediction horizon, comparing the different sets of predictions with Gaussian noise.}
    \label{fig:crpsgauss}
\end{figure}

Figure \ref{fig:crpsgauss} shows the CRPS (\ref{eq:crpsreal}) and its components, CRPS skill (\ref{eq:crpskill}) and CRPS spread (\ref{eq:crpspread}), for the selected Gaussian noise configurations. Values closer to zero indicate better performance of the set, as they reflect a balance between accuracy (low error of the predicted distribution) and adequate dispersion among the predictions of the set, which favors prediction sets that explore more alternatives with controlled error.

In this configuration, the CRPS is higher at the beginning of the forecast horizon when the first two entries correspond to noisy initial states. The figure shows that the CRPS skill curve reflects greater disagreement with actual observations in the early steps. Although the contribution of the CRPS spread helps to mitigate the error, it is not sufficient to achieve the values obtained with lower noise levels. In the latter cases, the lower disturbance of the initial conditions results in better performance in the first days of the forecast.

However, these configurations show a change in trend over the time horizon, reaching their worst performance at the end. This can be understood by analyzing the CRPS skill graph, where the differences in error between the configurations are minimal at the end of the time horizon. As their dispersion (CRPS spread) is lower, these errors are not corrected adequately. Therefore, although the dispersion between the members decreases slightly from the sixth day onwards, the model with the highest noise ends up being the one that best represents the predicted distribution compared to the actual observations.

\begin{figure}[ht!]
    \centering
    \includegraphics[width=0.9\linewidth]{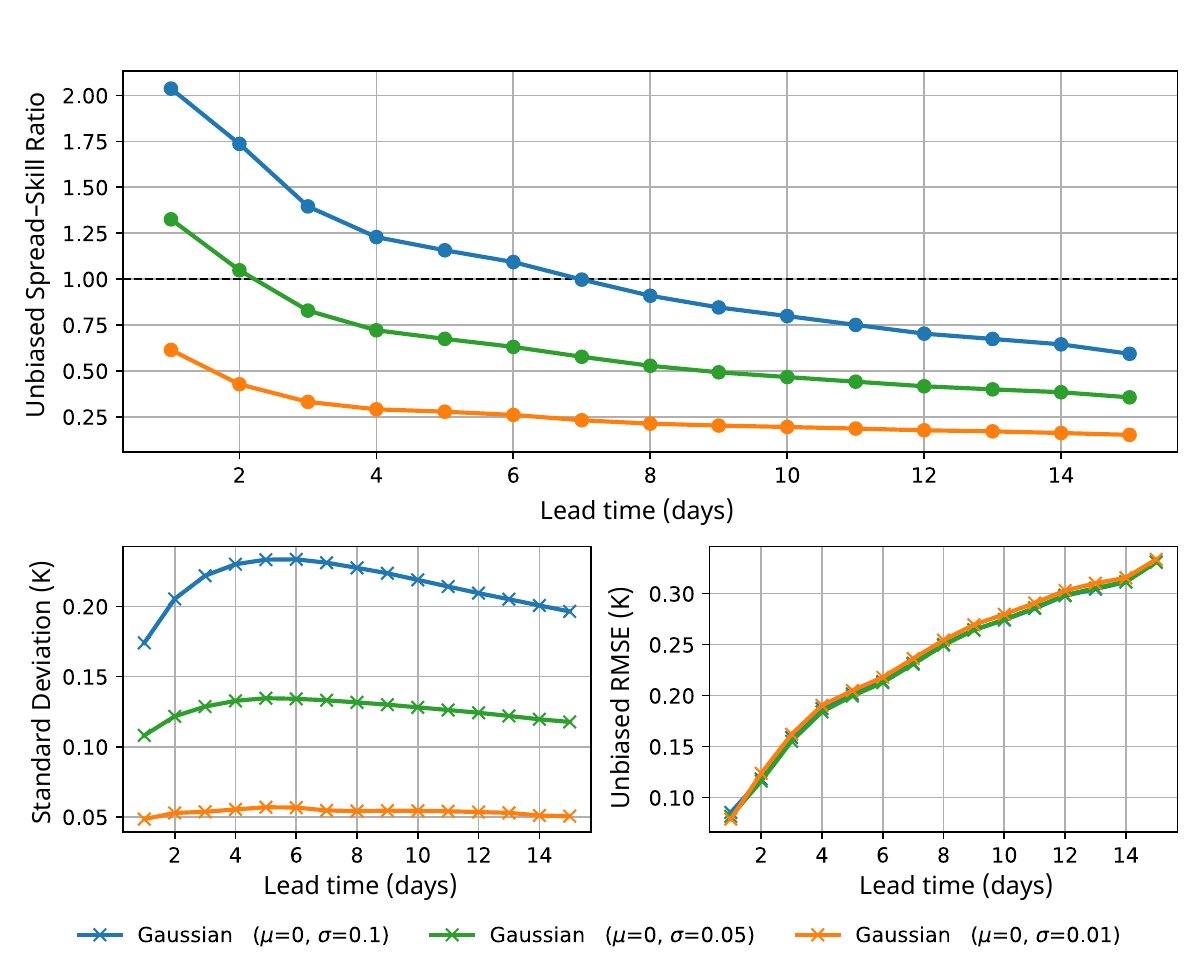}
    \caption{Graph of the unbiased spread-skill ratio and its components, averaged for each prediction horizon, comparing different sets of predictions with Gaussian noise introduced.}
    \label{fig:spreadskillgauss}
\end{figure}

Figure \ref{fig:spreadskillgauss} depicts the evolution of the spread-skill ratio (\ref{eq:r_values}) and its components: the standard deviation (\ref{eq:spread}) and the unbiased RMSE. The horizontal line at value 1 on the $y$-axis represents the reference threshold. This would be the case when the dispersion of the predictions of the set with respect to their mean can completely explain the errors of that mean, indicating perfect calibration.

In this case, the configuration with standard deviation equal to 0.01 presents a subdispersed behaviour throughout the entire prediction horizon, with insufficient dispersion with respect to the error committed by its mean. This can be verified by comparing the graph of standard deviation with respect to RMSE, where it is always a lower value. The other two configurations are overdispersed, with greater dispersion with respect to their mean. However, a common pattern is detected in the evolution of the spread-skill ratios, as they begin to stabilize from the sixth day onwards. This is consistent with the fact that the standard deviation of the sets grows until day 6, followed by a progressive decline. Furthermore, at this point it is also detected that the RMSE exceeds the standard deviation, which explains the coefficient less than one in the case of standard deviation equal to 0.1. It should be noted that this inflection point coincides with the moment when this configuration crosses the reference line, where the set would be perfectly calibrated. Therefore, although all configurations end up being underdispersed, it can be deduced that with a slightly larger but controlled disturbance, it would be possible to achieve a calibration of the set close to one at the end. 

\subsection{Comparison of Perlin noise configurations}
\begin{figure}[ht!]
    \centering
    \includegraphics[width=0.9\linewidth]{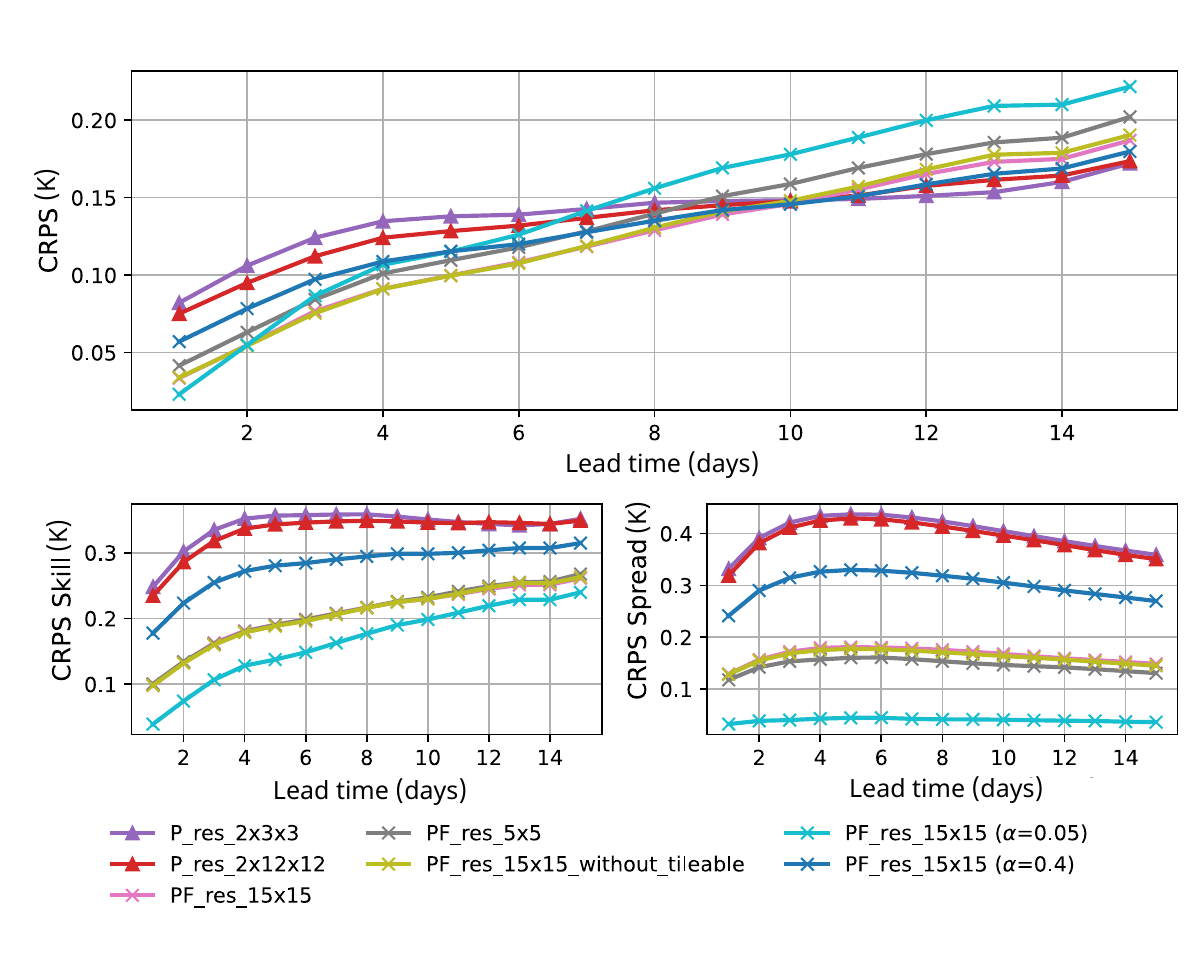}
    \caption{CRPS and its components, averaged for each prediction horizon, comparing the predictions with Perlin noise introduced.}
    \label{fig:crpsperlin}
\end{figure}

The Perlin noise results corresponding to the first days of prediction for CRPS are shown in Figure \ref{fig:crpsperlin}. The configurations with base Perlin noise (P\_res\_2x3x3 and P\_res\_2x12x12) show lower initial performance. However, these experiments end up being the best at the end of the time horizon compared to the fractal Perlin configurations, following a similar trend to Gaussian noise ($\mu$=0, $\sigma^2$=0.1). The configurations with Perlin fractal noise start with initial CRPS values close to the ideal, but their performance deteriorates over time.

One of the differences between the two groups is the complexity of the noise. Perlin fractal noise uses multiple iterations to refine the pattern, each with a higher resolution. However, the influence of these last octaves is minor, so only fine details are added. In contrast, base Perlin noise is simpler, as it generates the noise only once with the input resolution. Furthermore, fractal Perlin noise configurations incorporate a noise scalability factor, represented by $\alpha$, which is applied to the output to attenuate its impact.

The magnitude of the Perlin fractal noise with a scalability of 0.4 is comparable to the base Perlin noise. However, the dispersion between members is lower in this case, despite using a similar noise intensity. This suggests that fractal Perlin noise is not beneficial due to noise refinement through octaves, which generates complex spatial patterns with less structural coherence, thus affecting uncertainty.

\begin{figure}[ht!]
    \centering
    \includegraphics[width=0.9\linewidth]{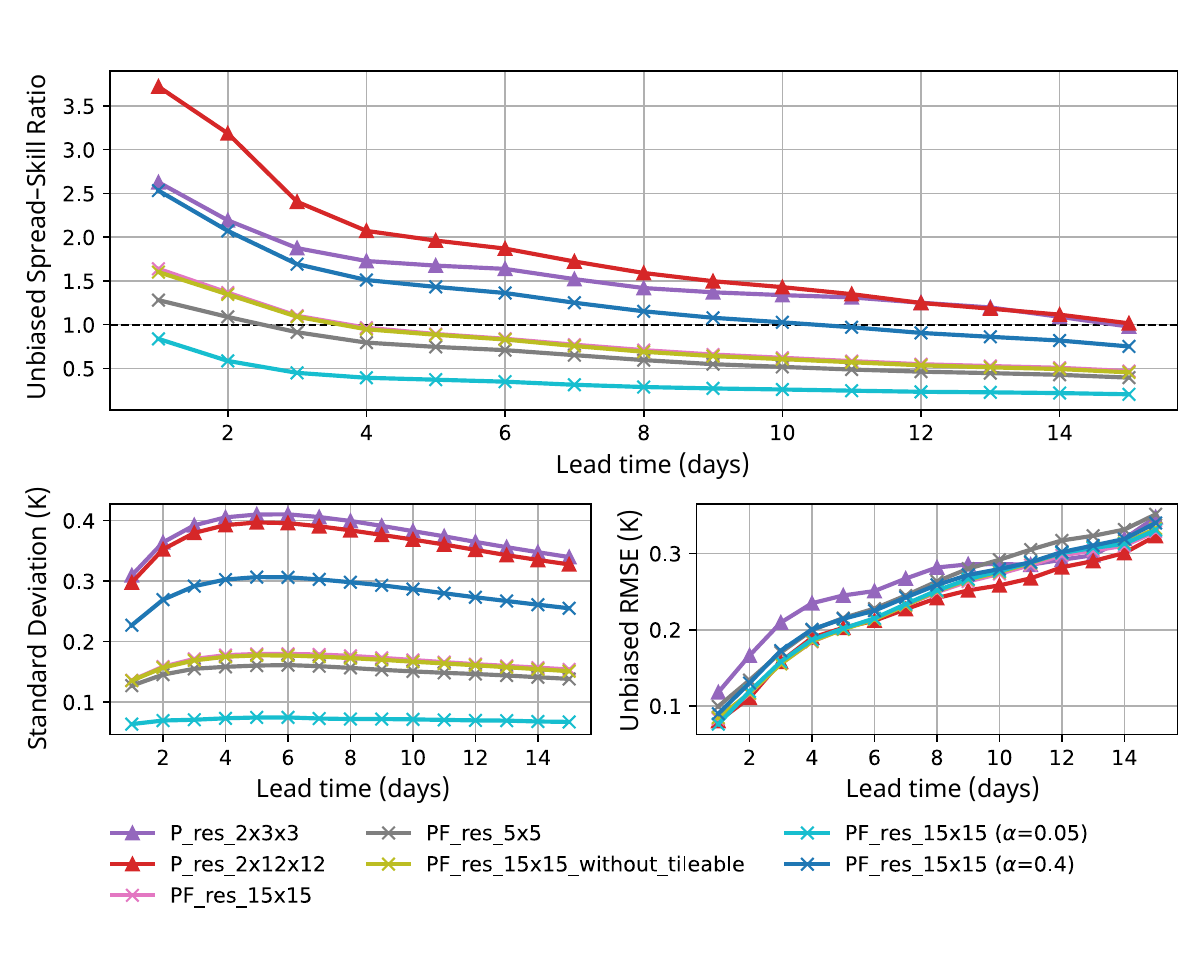}
    \caption{Unbiased spread-skill ratio and its components, averaged for each prediction horizon, comparing the different sets of predictions with Perlin noise introduced.}
    \label{fig:spreadskillperlin}
\end{figure}

Regarding the unbiased spread-kill ratio, Figure \ref{fig:spreadskillperlin} shows that the basic Perlin configurations stand out once again. Although they exhibit overdispersion over the time horizon, on the 15th day of prediction, they reach a value close to one, indicating an almost perfect calibration. Furthermore, it can be seen that in all sets, the standard deviation begins to decrease from the sixth day, as was already the case with Gaussian noise. This decrease, together with a less pronounced increase in unbiased RMSE, causes the spread-skill ratio to stabilize. On the contrary, due to the low intensity of the initial noise, most fractal Perlin configurations end up being underdispersed during most of the prediction period.

The P\_res\_2x3x3 configuration shares a very similar initial value, since, although its standard deviation is higher, its unbiased RMSE is also higher, especially at the beginning, compared to the other sets. This behavior of the unbiased RMSE highlights the importance of spatial resolution applied to noise, since the worst results tend to occur in configurations with lower resolutions, such as P\_res\_2x3x3 and PF\_res\_5x5.

\begin{figure}[ht!]
    \centering
    \includegraphics[width=1\linewidth]{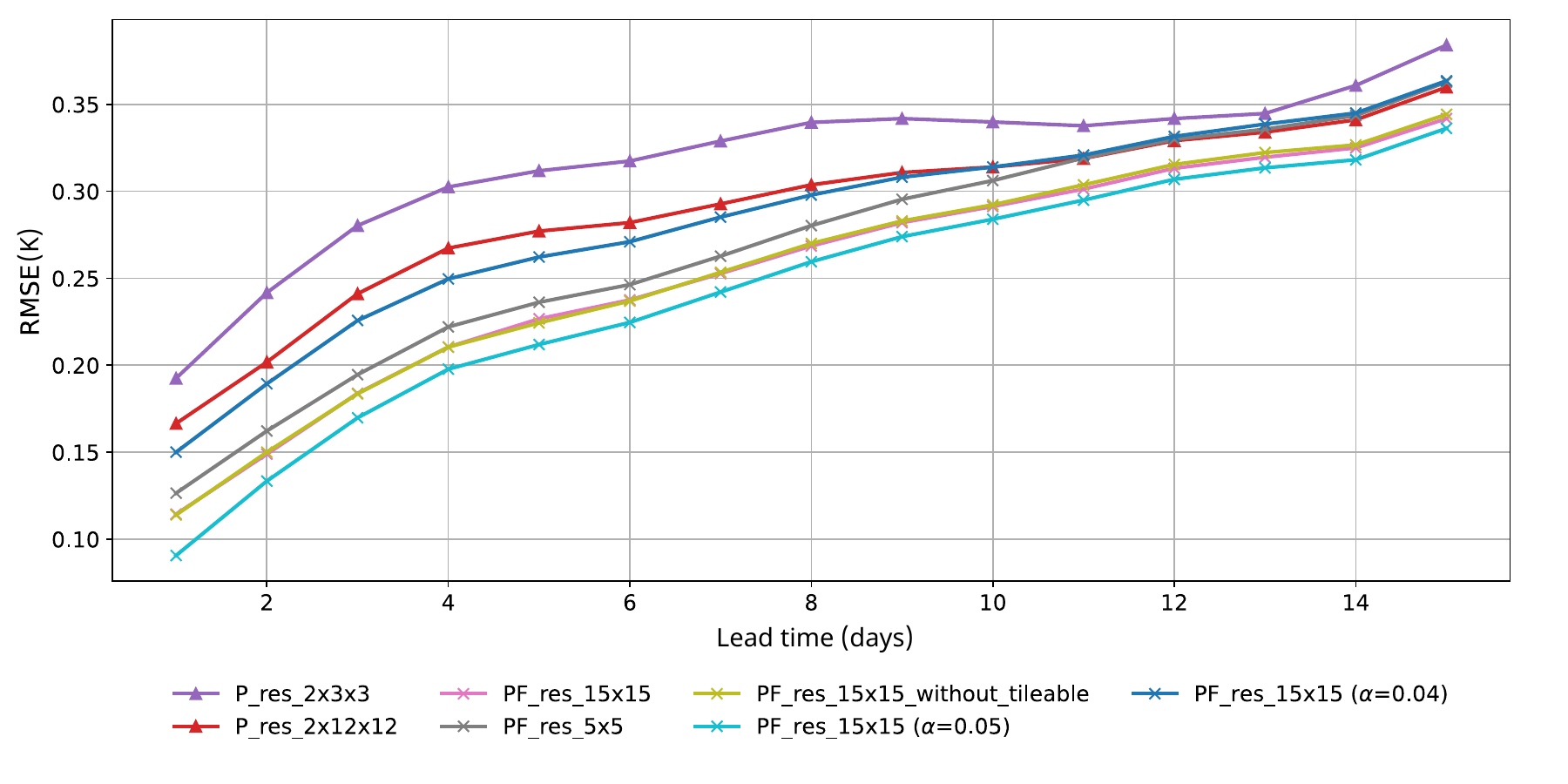}
    \caption{RMSE averaged by prediction horizon, comparing the performance of unifying Perlin noise configurations.}
    \label{fig:rmseperlin}
\end{figure}

Regarding the RMSE in the unification of predictions, Figure \ref{fig:rmseperlin} shows that higher noise again performs worse over the time horizon. However, the observation made with the unbiased RMSE is confirmed, since the resolution of the applied noise directly influences the model loss. Configurations with lower resolution generate fewer spatial patterns, which reduces the diversity of the set in different areas.

\subsection{Comparison between types of noise}

The most promising noises underwent an additional test in which predictions were made starting on the same day as the previous ones, but with the set size increased to 10. However, in this case, the results did not show significant differences, with minimal changes relative to prior trends. In the final evaluation, the noises with the best performance according to the previous metrics for each group analyzed were selected, along with one representative of each type of noise with the worst results for comparison purposes.

\begin{figure}[ht!]
    \centering
    \includegraphics[width=0.9\linewidth]{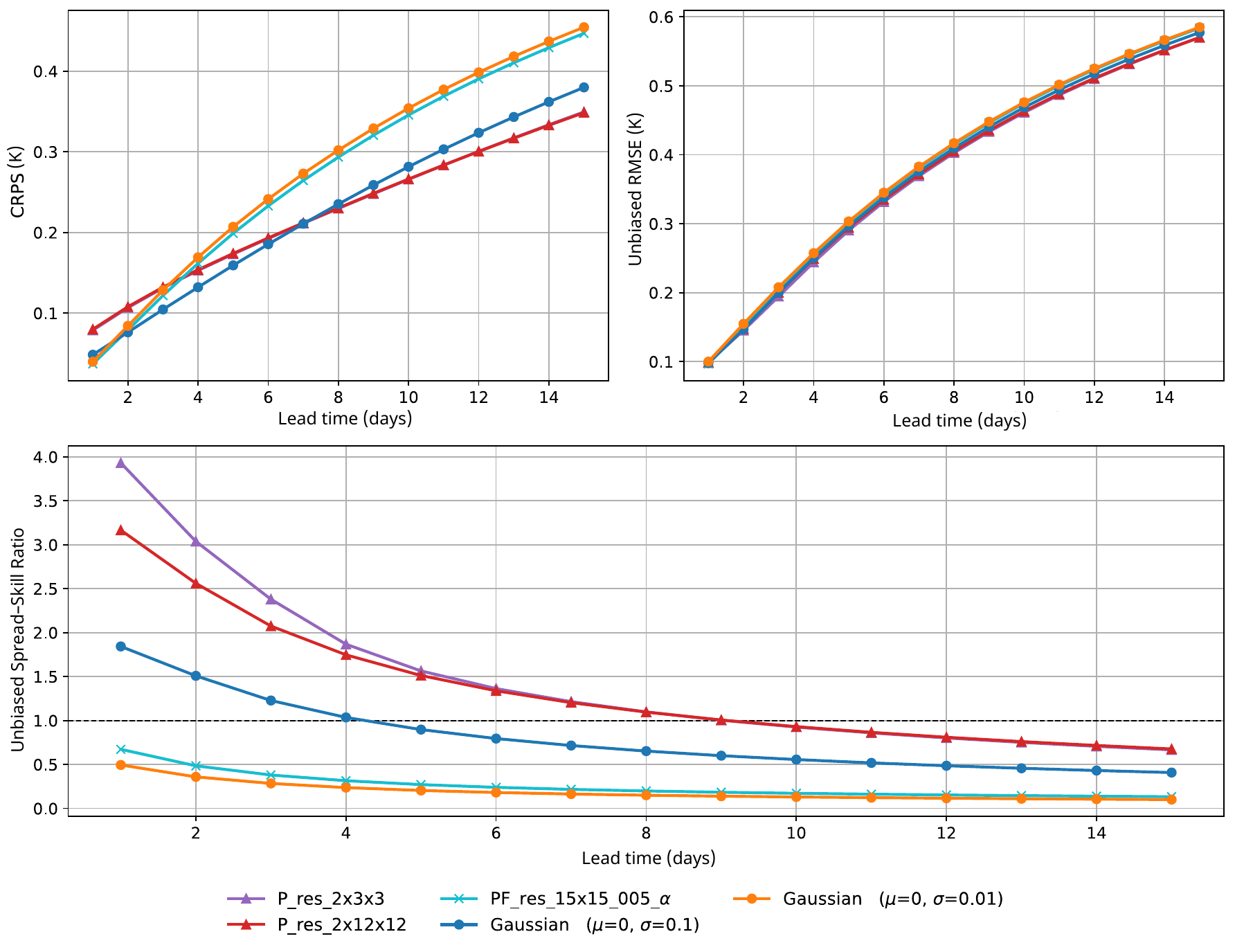}
    \caption{CRPS, unbiased RMSE, and spread-skill ratio averaged by prediction horizon, comparing the different sets of predictions with added noise, evaluated on the complete test data set.}
    \label{fig:ensprobeval}
\end{figure}

Figure \ref{fig:ensprobeval} corroborates the results observed in the preliminary tests, where the controlled intensity of the noise was a relevant factor for the set to be sufficiently diverse (for example, in the case of a Gaussian standard deviation of 0.1). Compared to those graphs, when we average over a large number of prediction start days, the metrics deteriorate overall, and the differences are smaller. Nevertheless, Perlin noise outperforms Gaussian noise configurations. 

Despite the difference in the statistical distribution of noise between the configurations of both groups, the noise intensities are similar in those with comparable performance (e.g., Gaussian noise $\sigma=0.1$ and Perlin base P\_res\_12x12). This suggests that the models benefit from initial perturbations that combine adequate noise intensity with some spatial structure.

\section{Discussion}
\label{se:discussion}

While deterministic error remains comparable to the single-model forecast, the predictions reveal important patterns in ensemble performance and uncertainty, providing a basis for further investigation.

The analysis of the initial perturbations reveals clear differences depending on their structure and intensity. Perturbations generated with moderate noise and structured spatial patterns (such as Perlin noise) proved more beneficial than those generated with scattered noise (e.g., Gaussian noise), which lack a clear spatial component. Furthermore, iterating over spatial noise patterns by introducing finer-scale details in successive steps (as in Perlin fractal noise) did not yield significant improvements. Among spatial patterns, higher-resolution configurations—characterized by a large number of spatial features—tended to produce better results. This may indicate that such perturbations enhance diversity at smaller spatial scales.

In addition to these findings, the characteristics of the base model provide important context for interpreting the results. The model used was relatively simple, both in terms of the number of variables and its training procedure, and it was trained deterministically. Despite this simplicity, several predictions generated from perturbed initial states achieved performance comparable to that of the noise-free model. This is a promising outcome, suggesting that large modified datasets, complex architectures, or highly sophisticated ensembles may not be strictly necessary to match the long-term performance of the base model. Although ensemble predictions tended to perform worse during the early stages, the diversity of errors among models appears to compensate over time, resulting in an average prediction similar to the deterministic one.

This study has identified several issues for further research beyond the exclusive perturbation of initial oceanographic data during inference. One possible direction is to train the model with autoregressive steps, which could improve the accuracy of long-term predictions. In this work, noise was added exclusively to the initial oceanographic states of the model. Considering that only one variable was used, this diversity may not be sufficient to capture complex phenomena such as coastal upwelling, which probably requires a greater number of variables to explain adequately. This scenario would also allow us to validate the conclusions drawn in this work regarding the types of noise that are most beneficial for creating larger-scale prediction sets.

Similarly, other types of ensemble learning techniques could be used for comparison, including the use of lagged predictions, the perturbation of initial forcing states, or the modification of the model's parameters. One of the advantages of ensemble learning techniques is their great flexibility, so future work could focus on introducing diversity in the training phase rather than inference, provided the necessary computational resources are available. Although no significant improvements were obtained in this work by increasing the number of ensemble members, this is likely due to the small number of predictions used, conditioned by computational limitations. This deficiency was evident in the analysis of the unbiased RMSE, where in several scenarios the results changed dramatically.

Therefore, this work aims to serve as an initial reference in the study of the types of noise applied during the inference phase as a mechanism for introducing diversity into prediction sets. However, as discussed throughout this section, the possibilities for configuring and incorporating diversity in this type of technique are extensive. As a result, future lines of research could focus on improving the results obtained in this work using the same techniques or exploring any of the other options mentioned.

\bibliographystyle{plain}
\bibliography{references}

\end{document}